\documentclass{article}
\usepackage{spconf,amsmath,graphicx}
\usepackage[shortlabels]{enumitem}

\usepackage{ifthen}
\usepackage{fancyhdr}
\usepackage{color}
\usepackage[table]{xcolor}
\usepackage[colorlinks=true,linkcolor=blue]{hyperref}

\title{GENERATIVE SCATTERNET HYBRID DEEP LEARNING (G-SHDL) NETWORK WITH
STRUCTURAL PRIORS FOR SEMANTIC IMAGE SEGMENTATION}
\name{Amarjot Singh and Nick Kingsbury}
\address{Signal Processing Group, Department of Engineering, University of Cambridge, U.K.}
\begin{document}

%
\maketitle
\begin{abstract}
This paper proposes a generative ScatterNet hybrid deep
learning (G-SHDL) network for semantic image segmentation. The proposed generative architecture is able to train
rapidly from relatively small labeled datasets using the introduced structural priors. In addition, the number of filters in each layer of the architecture is optimized resulting
in a computationally efficient architecture. The G-SHDL network produces state-of-the-art classification performance
against unsupervised and semi-supervised learning on two
image datasets. Advantages of the G-SHDL network over
supervised methods are demonstrated with experiments
performed on training datasets of reduced size.
\end{abstract}
\section{Introduction}
\label{sec:intro}
Semantic image segmentation is the task of partitioning and
labeling the image into pixel groups which belong to the
same object class. It has been widely used for numerous applications such as robotics~\cite{valada}, medical applications~\cite{singh2017}, augmented reality~\cite{miksik}, and automated driving~\cite{nadella}.

In the recent years, three types of learning architectures have been designed to learn the necessary representations required to solve the
semantic image segmentation task. These methods include
architectures that: (i) encode handcrafted features extracted
from the input images into rich non-hierarchical representations; (ii) learn multiple levels of feature hierarchies from the
input data; (iii) make use of the ideas from both categories to
extract feature hierarchies from hand-crafted features. 

He et al~\cite{he2016} is an example of the first class of architectures which utilize handcrafted region and global label features in multiscale conditional random fields to get the desired semantic segmentation. The second class of architectures includes Convolutional Neural Networks~\cite{long2014} and Deep Belief Networks~\cite{li2015} that learn multiple layers of features directly from the input images. These methods have been shown to
achieve state-of-the-art segmentation performance on various
datasets~\cite{garcia2017}. Despite their success, their design and optimal configuration is not well understood which makes it difficult
to develop them. In addition, the vast arrays of network parameters can
only be learned with the help of powerful computational resources and large training datasets. These may not be available
for many applications such as stock market prediction~\cite{jain2013},
medical imaging~\cite{singh2017} etc. The third class of models combine the concepts from both of the above-mentioned models to
learn shallow or deep feature hierarchies from low-level hand-crafted descriptors. Yu~\cite{yu2018} learned multiple layers
of hierarchical features from patch descriptors using stacked
denoising autoencoders. This class of models has produced
promising performance on various datasets~\cite{yu2018}.

This paper proposes the Generative ScatterNet Hybrid
Deep Learning (G-SHDL) network with structural priors for
semantic image segmentation. The G-SHDL network is inspired by the ScatterNet Hybrid Deep Learning (SHDL)~\cite{shdl2017}
network. The SHDL network extracts handcrafted features
from the input image using the ScatterNet front-end which are
then used by the unsupervised learning based Stacked PCA
mid-section layers to learn hierarchical features. These hierarchical features are finally used by the supervised back-end
module to solve the object classification task. The approximate minimization of the reconstruction loss function for the
PCA layers is obtained simply from the Eigen decomposition of the image patches~\cite{pcanet}. This results in rapid
learning of the hierarchical features. However we found that, despite the favorable
increase in the rate of learning, the approximate solution of
PCA loss function produces undesired checkerboard filters
which limit the performance of these models.

The proposed G-SHDL network is an improved version of
the SHDL network that uses ScatterNet as the front-end, similar to the SHDL network, to extract hand-crafted features
from the input images. However, instead of PCA layers in the
middle section, the G-SHDL uses four stacked layers of convolutional Restricted Boltzmann Machine (RBM) with structural priors to learn an invariant hierarchy of features. These
hierarchy features are finally used by a supervised conditional
random field (CRF) to solve the more complicated task of semantic segmentation as opposed to object recognition.

 \begin{figure*}[t!] 
\centering    
\includegraphics[width = 0.94\textwidth, height = 9.0 cm]{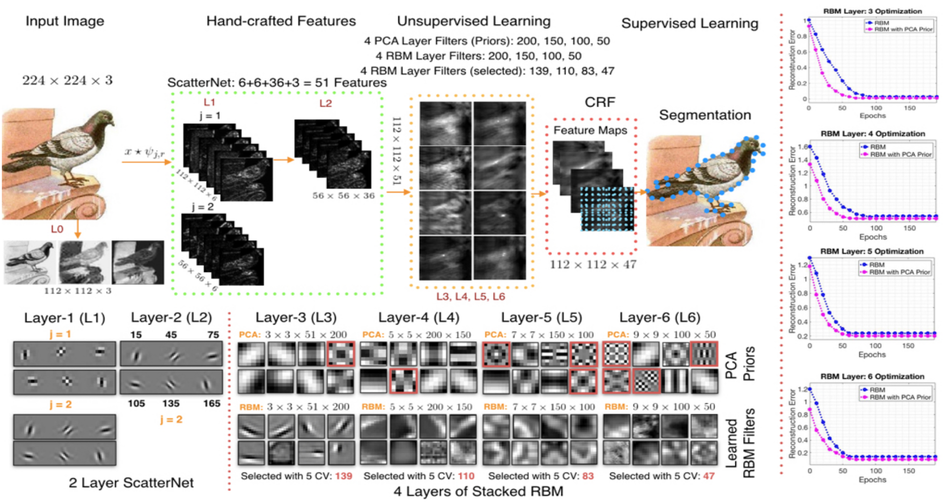}
\caption[Scattering operations performed on a signal to extract filtered response with translation invariance and to recover the lost frequency components. ]{\small{{The proposed G-SHDL network uses the ScatterNet front-end to extract hand-crafted scatternet features from the input image at L0, L1 and L2 using DTCWT filters at 2 scales and 6 fixed orientations (filters shown). The handcrafted features extracted
at the three layers are concatenated and given as input to the 4 stacked convolutional RBM layers (L3, L4, L5, L6) with 200, 150, 100 and
50 filters to learn a hierarchy of features. Each RBM layer is initialized with PCA based structural priors with same number of filters which improves their training as shown by L3 to L6 convergence graphs. The RBM layers are trained in a layer by layer greedy type fashion. Once
a RBM layer is trained the optimal number of filters are selected using 5 fold cross validation that results in a computationally efficient
architecture (Table. 1) as the later layers can feature from a smaller feature space. The features learned by the last RBM layer (L6) are used
by the CRF for semantic image segmentation. PCA layers can learn the undesired checkerboard filters (shown in red) which are avoided and
not used as the prior for the RBMs. In order to detect and remove the checkerboard filters from the learned filter set, we used the method defined in~\cite{geiger2012}.}}}
\label{fig:scatter00}
\end{figure*} 
  
The main contributions of the paper are stated below:

\begin{itemize}[topsep=2.2pt]
\itemsep0em 
\item \textit{Rapid Structural Prior based Learning of RBM}: Training of convolutional RBMs is slow as the partition function is approximated by sampling using MCMC (Section 2.2). In order to accelerate the training, the filters in each RBM layer are initialized with structural priors (filters) learned using PCA as opposed to random initialization. This has been shown to accelerate the training of RBMs (Fig. 1). Since, it is extremely fast to learn the filters or structural priors using PCA (eigen decomposition), the whole process is much faster than
training RBMs with random weight initialization.  
\vspace{-0.25em}
 
\item \textit{Computationally Efficient}:  The number of filters in
a particular RBM layer are optimized using crossvalidation that results in a computationally efficient
architecture as the filters in the subsequent layer are
now learned from a smaller feature space.
\vspace{-0.25em}
\item \textit{Advantages over supervised learning}: With G-SHDL only a fraction of the training samples need to be labelled, whereas supervised networks require large labelled training datsets for effective training, which may not be available~\cite{jain2013,yu2018}). The requirement for relatively small labeled datasets can be especially advantageous
for semantic segmentation tasks as it can be expensive
and time consuming to generate pixel-wise annotations.

  \end{itemize}

G-SHDL network is used to perform semantic segmentation on MSRC~\cite{shotton2006} and Stanford background (SB)~\cite{gould2009}
datasets. The average segmentation accuracy for each class
for both datasets is presented. In addition, an extensive comparison of the proposed pipeline with other deep supervised
segmentation methods is demonstrated.

The paper is divided as follows: section 2 briefly
presents the proposed G-SHDL network, section 3 presents
the experimental results while section 4 draws conclusions.

\vspace{-0.5em}

\section{PROPOSED G-SHDL NETWORK}
\label{sec:typestyle}
The Generative ScatterNet Hybrid Deep Learning Network (G-
SHDL) is detailed below. The first subsection explains the
mathematical formulation of the ScatterNet while the second
subsection presents the stacked RBM mid-section layers with
PCA structural priors that learn hierarchical features. The final sub-section explains the CRF supervised back-end that
uses the hierarchical features to produce the desired segmen-
tation. The G-SHDL network is presented in Fig. 1.
\vspace{-0.3em}
\subsection{DTCWT ScatterNet}
The parametric log based DTCWT ScatterNet~\cite{singhis2017} is used
to extract the relatively symmetric translation invariant hand-
crafted features from the RBG input image.

Invariant features are obtained by filtering the input
signal $x$ at the first layer (L1) with dual-tree complex wavelets~\cite{ngk,singhima} $\lambda_1 = (j,r)$ at different scales ($j$) and six pre-defined orientations
(r) fixed to $15^\circ, 45^\circ, 75^\circ, 105^\circ, 135^\circ$ and $165^\circ$. To build a
more translation invariant representation, a point-wise L 2
non-linearity (complex modulus) is applied to the real and
imaginary (a and b) of the filtered signal. The parametric log
transformation layer is then applied to all the oriented repre-
sentations extracted at the first scale $j = 1$ with a parameter
$k_{j=1}$, to reduce the effect of outliers by introducing relative
symmetry to the pdf~\cite{singhis2017}, as shown below 
   \begin{equation}
  U1[j] = \log(U[j] + k_{j}), \quad U[j] = \sqrt{|x\star \psi_{\lambda_{1} }^{a}|^2 + |x\star \psi_{\lambda_{1} }^{b}|^2},
\end{equation}

\begin{figure}[t!] 
\centering    
\includegraphics[width=1\linewidth, height = 3.00 cm]{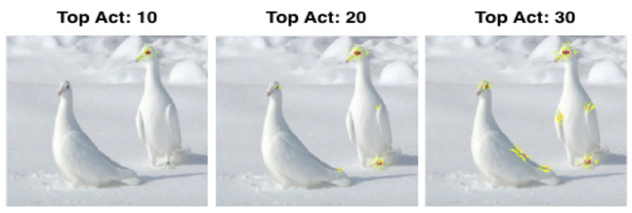}
\caption[Scattering operations performed on a signal to extract filtered response with translation invariance and to recover the lost frequency components. ]{\small{The illustration shows the L6 RBM features thresholded
to the top 10, 20 and 30 activations and back-projected to the input
pixel space~\cite{zeil}. The L6 RBM features are most responsive to the beaks of the birds, then feet and wings.}}
\label{fig:scatter00}
\end{figure}

Next, a local average is computed on the envelope $|U1[\lambda_{m = 1}]|$
that aggregates the coefficients to build the desired translation-
invariant representation:
\begin{equation}
S[\lambda_{m = 1}] = |U1[\lambda_{m = 1}]| \star \phi_{2^J}
\end{equation}

The high frequency components lost due to smoothing are retrieved by cascaded wavelet filtering performed at the second
layer ($L2$). The retrieved components are again not transla-
tion invariant so invariance is achieved by first applying the
L2 non-linearity to obtain the regular envelope followed by
a local-smoothing operator applied to the regular envelope
($U2[\lambda_{m = 1},\lambda_{m = 2}]$) to obtain the desired second layer ($L2$) coefficients with improved invariance:
\begin{equation}
S[\lambda_{m = 1},\lambda_{m = 2}] = |U1[\lambda_{m = 1}]| \star \psi_{\lambda_{2} }| \star \phi_{2^J}
\end{equation}
The scattering coefficients obtained at each layer are:
\begin{equation}
S = \begin{pmatrix}
x \star \phi_{2^J} (L0)\\
U1[\lambda_{m = 1}] \star \phi_{2^J} (L1)
\\ |U1[\lambda_{m = 1}]| \star \psi_{\lambda_{2} }| \star \phi_{2^J} (L2)
\end{pmatrix}_{j = (2,3,4,5...)}
\end{equation} 

ScatterNet features have been found to improve learning and generalization in deep supervised networks~\cite{singheff}.

\subsection{Unsupervised Learning: RBM with Priors}
The Scattering features extracted at (L0, L1, L2) are concate-
nated and given as input to 4 stacked convolutional restricted
Boltzmann machine (RBM) layers that learn 200, 150, 100 and 50 filters respectively. The RBM is a generative stochas-
tic neural network that learns a probability distribution over
the scattering features. Markov chain Monte Carlo (MCMC)
sampling in the form of Gibbs sampling is used to approximate the likelihood and its gradient. The estimation of
the likelihood of the RBM or its gradient for inference is computationally intensive~\cite{montavon}. However, initializing RBMs with
priors on the hidden layer instead of a random initialization
has been shown to improve the training~\cite{montavon}.

We propose structural priors for each convolutional RBM
layer (L3 to L6) which have been shown to improve the training of
the RBMs (Fig. 1 Graphs). The Structural priors are obtained
using the PCANet~\cite{pcanet} layer that learns a family of orthonormal filters by minimizing the following reconstruction error:
\begin{equation}
\min_{V \epsilon \ R^{z_{1}z_{2}\times K} } \left \|X-VV^TX  \right \|_{F}^2,\ s.t.\  VV^T = I_{K}
\end{equation} 
where $X$ are patches sampled from $N$ training images (concatenated handcrafted features), $I_K$ is an identity matrix of size
$K \times K$. The solution of eq. 5 in its simplified form represents K leading principal eigenvectors of XX T obtained
using Eigen decomposition. The PCA layers may learn
undesired checkerboard filters. In order to detect the checker-board filters from the learned filter set, we use the method
defined in~\cite{geiger2012}. These checkerboard filters are avoided as
filter priors. Each RBM layer (L3, L4, L5, L6) of the G-SHDL is trained individually in a greedy fashion (with structural priors). Once the RBM layer is trained the filters that
learn redundant information are removed using 5 fold cross-validation. (Table 1 and section 3.2).
\begin{figure}[t!] 
\centering    
\includegraphics[width=0.96\linewidth, height = 3.25 cm]{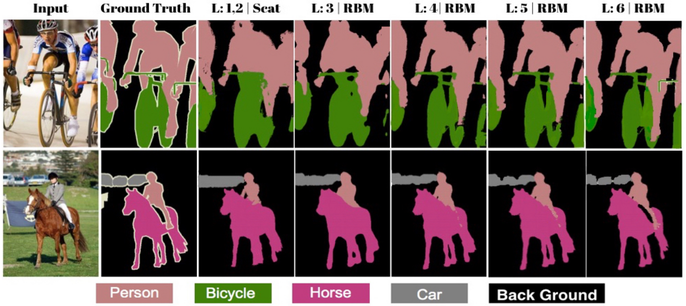}
\caption[Scattering operations performed on a signal to extract filtered response with translation invariance and to recover the lost frequency components. ]{\small{Figure shows two images from MSRC dataset with their
ground truth and segmentation obtained at L2 to L6 of G-SHDL.}}
\label{fig:scatter00}
\end{figure}

\subsection{Supervised CRF Segmentation}
Conditional Random Field (CRF) is a probabilistic graphical model that uses the features obtained from the L6 RBM along with edge potentials computed on 4 pairwise connected grids~\cite{domke} to perform the desired segmentation. The segmentation is obtained by minimizing the clique loss function with Tree-Reweighted~\cite{domke} inference that uses the LBFGS optimization algorithm.
\vspace{-2em}
\section{Overview of Results}
\label{headings}
G-SHDL was evaluated and compared with other segmentation frameworks on both MSRC~\cite{shotton2006} and Stanford Background (SB)~\cite{gould2009} datasets. The MSRC dataset contains 591 images with 21 classes while the SB dataset is formed of 715
images with 8 classes, where each image in both datasets has a resolution of 320$\times$240. The quantitative results are presented with the class pixel accuracy which represents the ratio of correct pixels computed in a per-class (PA)~\cite{garcia2017} basis and then averaged over the total number of classes. The results are presented for 5-fold cross-validation for both datasets randomly split into 45\% training, 15\% validation and 40\% test images
for each fold. We provide a quantitative comparison against the state-of-the-art to evaluate the performance of G-SHDL.

\subsection{Handcrafted Front-end: ScatterNet}
ScatterNet features are extracted from the input RGB image using DTCWT filters at 2 scales and 6 fixed orientations. Next, log transformation with parameter $k_{j=1}$ = 1.1 is applied to the representations obtained at the finer scale to introduce
relative symmetry. (Section. 2.1).
\vspace{-0.3em}
\begin{table}[!t]
\centering
\caption{\small{5 fold cross validation performed on the training dataset of
Stanford background (SB) dataset to select optimal filters for L3 to
L6 RBM layers. L(size) = No. of Filters ($a,a$ is equivalent to $a \times a$)}}
\label{MNIST}
\begin{tabular}{c|cccc}
\hline
\small{Filters} & \small{L3 (size)} & \small{43 (size)} & \small{L5 (size)} & \small{L6 (size)}\\
 \hline
\small{PCA}  & 200 (3,3) & 150 (5,5) & 100 (7,7) & 50 (9,9)  \\
\small{RBM} & 200 (3,3) & 150 (5,5) & 100 (7,7) & 50 (9,9) \\ \hline
\small{Selected} & 139 & 110 & 83 & 47\\
\end{tabular}
\end{table}
\vspace{-0.3em}
\subsection{Unsupervised Mid-section: RBM with PCA priors}
The four stacked convolutional RBM layers learn 200, 150,
100 and 50 filters respectively with PCA structural priors (obtained by training on the handcrafted features) in a greedy
layer-wise fashion (Section 2.2). Once, each RBM layer is
trained, five-fold cross-validation (5-CV) is computed with
filters randomly selected from the trained filter set to evaluate the segmentation accuracies using CRF. We are able to
achieve similar PA accuracy on the 5-CV with the fewer number of filters than the complete learned filter set. This suggests
that some of the filters learn redundant information which can
be removed. This results in efficient learning of subsequent
layers as the filters are learned from a smaller feature space.
The numbers of selected filters are shown in Table. 1.
\vspace{-0.3em}
\subsection{Classification performance and comparison}
This section presents the classification performance of each
module of the G-SHDL network. The accuracy of the hand-crafted module (HC) is computed on the concatenated relatively symmetric features extracted at L0, L1, L2, for both resolutions (R1, R2) using CRF for segmentation on MSRC dataset. The hand-crafted module produced a classification accuracy of 68.4\% (HC) as shown in Table. 2. An increase
of approximate 4\%, 2\%, 2\% and 2\% is observed when the
mid-level features, learned at L3, L4, L5 and L6 are used by
the CRF. This suggests that the RBM layers learn useful image representations as they improve the segmentation performance finally producing an accuracy of 78.21\%.
\vspace{-0.3em}
\begin{table}[!h]
\centering
\caption{\small{PA (\%) on SB dataset for each module computed with CRF. The increase in accuracy with the addition of each layer is also
shown. HC: Hand-crafted. RBM Layers: L3, L4, L5 and L6.}}
\label{MNIST}
\begin{tabular}{c|ccccc}
\hline
 Dataset & HC & L3 & L4 & L5 & L6\\
 \hline
 Accuracy & 68.4 & 72.3 & 74.8 & 76.7 & \cellcolor{gray!50}\textbf{78.21}
\end{tabular}
\end{table}

Next, the performance of the SHDL network is evaluated
on the MSRC dataset. The network results in a segmenta-
tion accuracy of 83.90\%, as shown in Table. 3. The G-SHDL outperformed the semi-supervised and unsupervised
learning methods on both datasets; however the network underperformed against supervised deep learning models~\cite{jin,liu}, as shown in Table 3. The segmentation results for two
images from the MSRC dataset are shown in Fig. 3.
\vspace{-1.2em}
\begin{table}[!h]
\centering
\caption{\small{PA (\%) and comparison on both datasets. Unsup: Unsu-
pervised, Semi: Semi-supervised and Sup: Supervised.}}
\label{MNIST}
\begin{tabular}{c|ccc|c}
\hline
 \small{Dataset} & \small{G-SHDL} & \small{Semi} & \small{Unsup} & \small{Sup}\\
 \hline
\small{SB}~\cite{shotton2006} & \cellcolor{gray!50}78.21 & 77.76~\cite{souly} & 68.1~\cite{coates} & \textbf{80.2}~\cite{collobert}\\
\small{MSRC}~\cite{gould2009} & \cellcolor{gray!50}83.90 & 83.6~\cite{liu2013} & 74.7~\cite{rubinstein} & \textbf{89.0}~\cite{liu}\\

\end{tabular}
\end{table}
\vspace{-1em}
\subsection{Advantage over Deep Supervised Networks}
Deep Supervised models need large labeled datasets for training which may not exist for most application. Table 4 shows that our G-SHDL network outperformed the recurrent
CNN of~\cite{collobert} on the SB dataset with less than 300 images due
to poor ability of rCNNs to train on small training datasets. The
experiments were performed by dividing the training dataset
into 8 datasets of different sizes. It is made sure that an equal
number of images per object class were sampled from the training dataset. The full test set was used for all experiment.
\vspace{-5mm}
\begin{table}[!h]
\centering
\caption{\small{Comparison of G-SHDL on PA (\%) with Recurrent CNN
(rCNN)~\cite{collobert} against different training dataset sizes on SB dataset.}}
\label{MNIST}

\begin{tabular}{c|ccccccc}
\hline
\small{Arch.} & \small{50} & \small{100} & \small{200} & \small{300} & \small{400} & \small{500} & \small{572}\\
 \hline
\small{G-SHDL} & \cellcolor{gray!50}\textbf{\small{40.3}} & \cellcolor{gray!50}\textbf{\small{59.9}} & \cellcolor{gray!50}\textbf{\small{66.4}} & \cellcolor{gray!50}\textbf{ \small{72.6}} & \small{75.7} & \small{78.20} & \small{78.21}\\
\small{rCNN} & \small{15.6} & \small{34.5} & \small{41.1} & \small{66.9} & \cellcolor{gray!50}\textbf{\small{76.2}} & \cellcolor{gray!50}\textbf{\small{79.87}} & \cellcolor{gray!50}\textbf{\small{80.2}}\\
\end{tabular}
\end{table}
\vspace{-1.3em}
\section{Conclusion}
The paper proposes a generative G-SHDL network for semantic image segmentation that is faster to train and computationally efficient. The network uses PCA based structural priors that accerlate the training of (otherwise slow)
RBMs. The network has been shown to outperform unsupervised and semi-supervised learning methods while evidence
of the advantage of G-SHDL network over supervised learning (rCNN) methods is presented for small training datasets.
\bibliographystyle{IEEEbib}
\bibliography{refs}

\end{document}